%% file: main.tex
\documentclass[letterpaper]{article} 
\usepackage{aaai2026}
\usepackage{times}  
\usepackage{helvet}  
\usepackage{courier}  
\usepackage[hyphens]{url}  
\usepackage{graphicx} 
\usepackage{natbib}  
\usepackage{caption} 
\usepackage{algorithm}
\usepackage{algorithmic}
\usepackage{amsmath} 
\usepackage{amssymb}  

\usepackage{arydshln}
\usepackage[table]{xcolor}

\usepackage{newfloat}
\usepackage{listings}
\DeclareCaptionStyle{ruled}{labelfont=normalfont,labelsep=colon,strut=off} 
\floatstyle{ruled}
\newfloat{listing}{tb}{lst}{}
\floatname{listing}{Listing}
\usepackage{booktabs}

\title{Clinician-in-the-Loop Smart Home System to Detect Urinary Tract Infection Flare-Ups via Uncertainty-Aware Decision Support}
\author{
    Chibuike E. Ugwu\textsuperscript{\rm 1},
    Roschelle Fritz \textsuperscript{\rm 2},
    Diane J. Cook\textsuperscript{\rm 1},
    Janardhan Rao Doppa\textsuperscript{\rm 1}
    }
\affiliations{
    \textsuperscript{\rm 1}School of Electrical Engineering and Computer Science, Washington State University, USA\\

    \textsuperscript{\rm 2} University of California, Davis Health, Betty Irene Moore School of Nursing\\
}

\usepackage{bibentry}

\begin{document}

\maketitle
\begin{abstract}
Urinary tract infection (UTI) flare-ups pose a significant health risk for older adults with chronic conditions. These infections often go unnoticed until they become severe, making early detection through innovative smart home technologies crucial. Traditional machine learning (ML) approaches relying on simple binary classification for UTI detection offer limited utility to nurses and practitioners as they lack insight into prediction uncertainty, hindering informed clinical decision-making. This paper presents a clinician-in-the-loop (CIL) smart home system that leverages ambient sensor data to extract meaningful behavioral markers, train robust predictive ML models, and calibrate them to enable uncertainty-aware decision support. The system incorporates a statistically valid uncertainty quantification method called Conformal-Calibrated Interval (CCI), which quantifies uncertainty and abstains from making predictions (``I don’t know") when the ML model's confidence is low. Evaluated on real-world data from eight smart homes, our method outperforms baseline methods in recall and other classification metrics while maintaining the lowest abstention proportion and interval width. A survey of 42 nurses confirms that our system's outputs are valuable for guiding clinical decision-making, underscoring their practical utility in improving informed decisions and effectively managing UTIs and other condition flare-ups in older adults.


\end{abstract}


\section{Introduction}
With the aging of the global population, individuals, families, and communities face the growing challenge of caring for older adults who are diagnosed with multiple chronic health conditions. In 2023, 78.8\% of older adults reported multiple chronic conditions \cite{watson2025trends}. Chronic conditions account for 90\% of the amount the US spends on healthcare yearly \cite{usdhhs_multiple_chronic_conditions}. Living with chronic conditions is linked to greater functional decline, disability, diminished quality of life, higher likelihood of hospitalization, and increased risk of mortality \cite{newman2020correction}.

Effective management of chronic health conditions through partnerships with home health nurses can significantly reduce hospital use and care costs while improving quality of life \cite{sylvia2008guided}. However, the complexity of disease management and the rapidly decreasing ratio of healthcare professionals to patients are putting demands on health management that cannot be met through traditional mechanisms \cite{ekstedt2023patient}. Until more effective treatments are available to halt or reverse many chronic conditions, technology—such as smart home monitoring systems—will be an essential tool for bridging the gap between growing health needs and limited care accessibility. 

Smart homes suggest a way to improve management of chronic health conditions and scale nurse capabilities by monitoring behavior patterns and using the information to detect and report condition flare-ups. Currently, to address a flare-up of a chronic condition, caregivers and clinicians rely on inconsistent and unclear client self-reports or very brief in-person assessments, challenging effective condition management \cite{taniguchi2020visiting}. The resulting inadequacy of at-home management and lack of awareness of condition flare-ups represent leading causes of condition-related hospitalization \cite{fan2020remote}. Smart homes lay the groundwork for automating behavior and health monitoring, which can supplement clinical visits, scale the reach of nurses with limited bandwidth, and improve the effectiveness of healthcare systems. By monitoring behavior patterns and anomalies, researchers have used data collected from ambient sensors in smart homes and other environments to detect cognitive decline \cite{robben2016delta}, assess sleep quality \cite{hasan2024sleep}, diagnose health conditions \cite{sprint2024building}, and sense emergency conditions such as falls \cite{vaiyapuri2021internet}. 

Despite their successes, smart homes have limitations for managing chronic health conditions. First, the variability of human behavior for even healthy individuals complicates the task of modeling behavior from ambient sensor data. Second, health conditions can impact people in different ways, particularly when they occur in combination. Third, current smart home technology focuses on analyzing and reporting detected situations, rather than acting on them. To harness the power of these technologies, smart homes can partner with clinicians to provide a complete cycle of sensing, detecting, identifying, and acting. We refer to such a system as a clinician-in-the-loop (CIL) smart home.

In such a CIL system, the reliability of the system’s predictions becomes critical. Clinicians often need to decide whether to escalate care, adjust treatment, or monitor further based on smart home alerts. If the system’s output comes without a clear measure of confidence, there is a risk of both false alarms (leading to unnecessary interventions) and missed detections (leading to adverse health outcomes).

Uncertainty quantification (UQ) addresses this gap by providing a principled way to express how confident the system is in its classifications or predictions. In the context of chronic condition management, UQ can help prioritize alerts that are highly reliable, flag borderline cases for closer review, and support clinicians in balancing sensitivity with specificity. Moreover, when UQ methods come with statistical guarantees—such as those offered by conformal prediction—clinicians gain an additional layer of trust, knowing that reported uncertainty levels have rigorous backing.

We hypothesize that clinicians can act more confidently and effectively on smart home-detected condition flare-ups when the technology provides clear information on prediction uncertainty. Additionally, clinicians are more likely to trust and adopt the technology if the uncertainty in its predictions is accompanied by reliable guarantees. The main contributions of this paper are:
\begin{itemize}
    \item Design a clinician-in-the-loop smart home framework that extracts clinically relevant behavior markers from ambient sensor data to predict UTI flare-ups.
    \item Introduce a novel Conformal-Calibrated Interval (CCI) method to quantify predictive uncertainty, with statistically valid coverage guarantees, and enable abstention from making predictions when confidence is low.
    \item Demonstrate empirically, on 117 labeled days from eight smart homes, that CCI produces compact intervals, achieves higher recall, and reduces abstention rates compared to baseline methods.
    \item Validate our approach through a survey of 42 nurses, to show that CCI's outputs improve interpretability and support clinical decision-making in managing UTI flare-ups.
\end{itemize}

\begin{figure}
    \centering
    \includegraphics[width=1\linewidth]{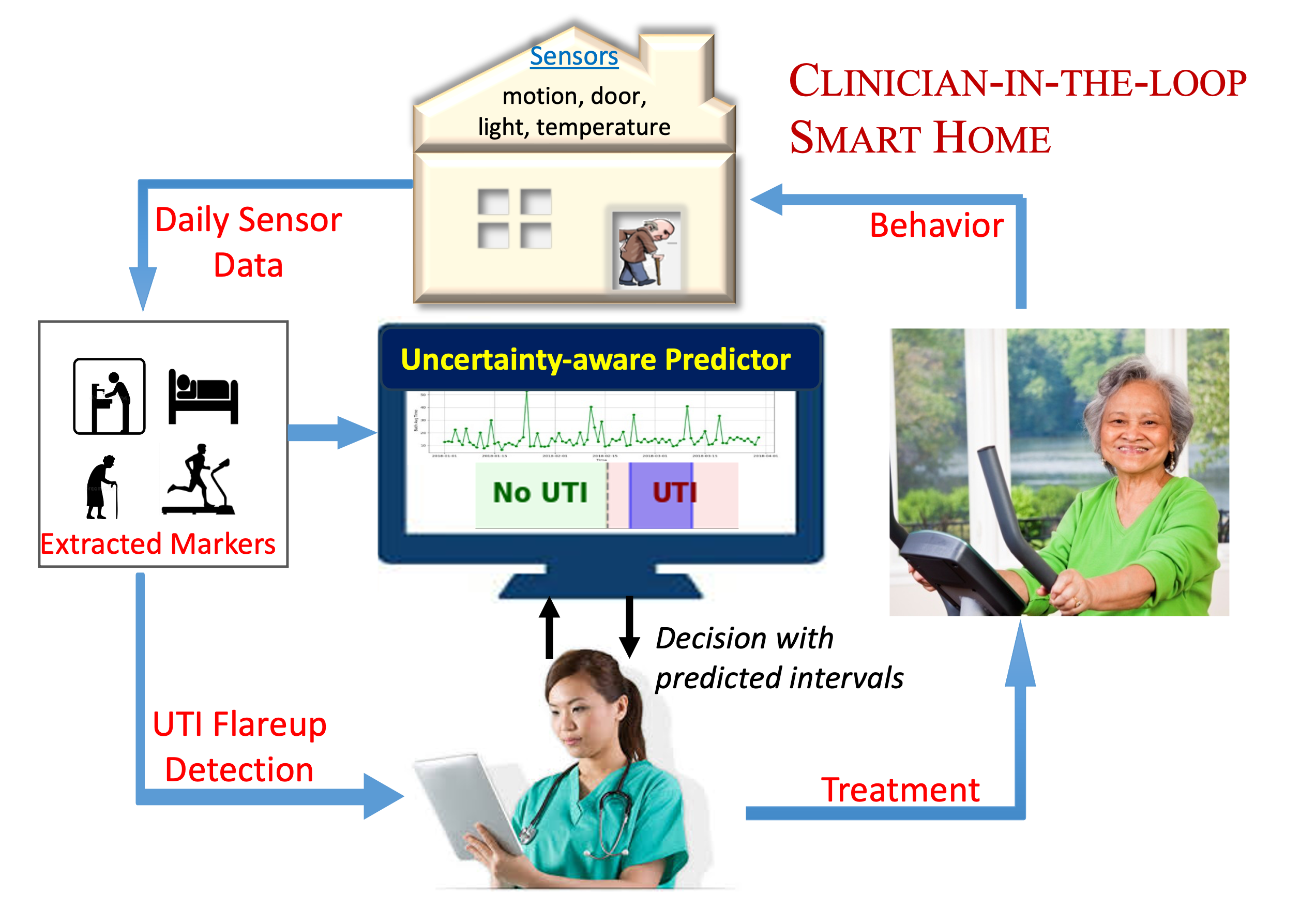}
    \caption{Overview of the clinician-in-the-loop smart home system for UTI flare-up detection. The system analyzes daily sensor data to extract behavioral markers and predict UTI occurrence in an uncertainty-aware manner. Clinicians, integrated into the loop, use these prediction/uncertainty intervals to guide decision-making and treatment planning.}
    \label{fig:UTI_main_image}
\end{figure}

\section{Related Work}
Prior efforts to detect urinary tract infections (UTIs) using smart home data include the work by Capstick et al.~\cite{capstick2024digital}, who developed a machine learning (ML) algorithm for early UTI screening in people living with dementia. Their approach utilized low-cost passive sensors to monitor in-home activity and physiological patterns. The algorithm focuses on extracting features and generating point predictions to alert clinicians. Complementary research by \cite{jeng2022machine} developed and validated ML models (logistic regression, decision tree, and random forest) algorithms to predict recurrent UTIs. Their models integrated patient characteristics with bacterial features across both initial clinical visits and post-hospitalization stages. However, the approach does not provide information about model uncertainty or offer theoretically-grounded guarantees on performance, such as coverage probabilities for prediction sets/intervals. In contrast, {\em we address these limitations by incorporating UQ and providing distribution-free theoretical guarantees on the reliability of UTI detections.}

Conformal prediction (CP) is a distribution-free framework for uncertainty quantification that produces prediction sets or intervals with finite-sample coverage guarantees under the mild assumption of exchangeable data~\cite{vovk2005algorithmic, shafer2008tutorial, angelopoulos2023conformal, romano2019conformalized, NCP, PRCP, R3CP, LLM-CP, DPSM, IJCAI-2025, DAC-2025}. Unlike traditional probabilistic models that rely on strong distributional assumptions, CP provides calibrated uncertainty estimates regardless of the underlying model, making it well-suited for safety-critical applications. {\em CP is widely used in regression and multi-class classification, but we adapt it here to binary classification for UTI detection}, constructing valid prediction intervals around the predicted probabilities to enable uncertainty-aware decisions rather than relying solely on point probability predictions.

\section{Background and Problem Setup}

\subsection{UTI Detection from CIL smart homes}
We employ data collected in the homes of older adults who are managing multiple chronic health conditions. We installed ambient sensors in the homes that include passive infrared motion detectors, magnetic door detectors, and sensors for ambient light and temperature \cite{cook2012casas}. We collected continuous data for one year in each home while participants performed their normal daily activities. Chronic conditions included congestive heart failure, chronic obstructive pulmonary disorder, diabetes, cancer, irritable bowel syndrome, diverticulitis, atrial fibrillation, arthritis, epilepsy, ulcerative colitis, restless leg syndrome, Parkinson’s disease, and Sjogren’s syndrome. The study was approved by the Washington State University Institutional Review Board.

In this paper, we focus on detecting urinary tract infections (UTIs) from smart home data in an uncertainty-aware manner. UTIs are among the most common bacterial infections worldwide \cite{mancuso2023urinary}. We selected this class of health event based on its prevalence in our study and the critical need for detection and treatment. Many older patients with UTIs fail to recognize symptoms before the infection becomes damaging and they are sent to an emergency room \cite{rodriguez2020urinary}.

UTI symptoms are atypical in older adults. While they often present with painful urination and increasing urinary frequency, other symptoms may include delirium, confusion, dizziness, drowsiness, and falls. The goal of this AI application is to detect UTIs in the early stages so they can be more quickly and effectively treated. Little work has been done to automatically detect this condition, so clinicians typically rely on laboratory testing once an infection is suspected.

\subsection{Data Collection}
Weekly telehealth visits were conducted by registered nurses throughout the study to obtain ground truth labels. During these visits, participants reported changes in health status, including UTI flare-ups (noted in eight homes). Nurses documented their observations, participant symptoms, interview summaries, and vital signs. Based on these records, nurses annotated specific days with UTI flare-ups, forming the UTI-positive labels in our dataset. To create a balanced and contextually relevant dataset for model development and evaluation, all UTI-positive days were included, along with adjacent UTI-negative days (days immediately preceding or following a flare-up but without a UTI). 

Our dataset includes 117 samples from eight participants, each with at least one documented UTI event during the study. Participant demographics and sample counts are summarized in Table~\ref{table:participant_demographics}. Across all participants, the dataset includes 117 samples, with a mean age of 83.8 years (SD = 15.8), five females and three males, an average of 15.8 years of education (SD = 3.3), and 56 UTI vs. 61 non-UTI days.

\begin{table}[h]
    \centering
    \begin{tabular}{|c|cccc|}
        \hline
        \textbf{Participant} & \textbf{} & & \textbf{Education} & \textbf{No.} \\
        \textbf{ID} & \textbf{Age} & \textbf{Gender} & \textbf{(years)} & \textbf{Samples} \\
        \hline
        P1   & 83  & Male     & 16  & 38 \\
        P2   & 88  & Female   & 14  & 13 \\
        P3   & 89  & Male     & 18  & 15 \\
        P4   & 86  & Female   & 12  & 10 \\
        P5   & 90  & Male     & 21  & 9  \\
        P6   & 73  & Female   & 14  & 13 \\
        P7  & 78  & Female   & --  & 9  \\
        P8  & 83  & Female   & --  & 10 \\
        \hline
        Total & & & & 117 \\
        \hline
    \end{tabular}
    \caption{Demographic information and number of samples from each participant.}
    \label{table:participant_demographics}
\end{table}

\subsection{Behavioral Feature Analysis}
From continuous ambient smart home sensor data, we extracted 5 clinically relevant behavioral features associated with potential UTI flare-ups. These were selected from an initial set of 17 extracted features: \textit{Nocturnal Bathroom Visits, Nocturnal Non-Bathroom Movement, Percentage of Nocturnal Bathroom Visits, Health Event (last 3 days), and Daily Movement Entropy}. A detailed description of all 17 features is in the Appendix. Feature importance was assessed using SHAP values to identify the top five-ranked features, upon which our main results are based. 

\section{Methodology: Generation of CCI Intervals}
In this section, we present two methods for uncertainty-aware decision-making for detecting UTIs. Specifically, we describe: Naive Intervals, which generate simple uncertainty intervals; and our proposed method, Conformal-Calibrated Intervals, which deliver statistically valid and easily interpretable prediction intervals through conformal prediction.

\subsection{Naive Uncertainty Intervals}

As a simple baseline for quantifying uncertainty in the model's predictions, we construct \textit{Naive Intervals} using the raw ensemble output from a Random Forest classifier because of its stability compared to other ensembles (e.g., neural network). Each decision tree in the forest outputs a probability estimate $p_j(x)$ for the positive class (UTI), and the final predicted probability is the average $\hat{p}(x)$ across all trees:
To capture uncertainty in this estimate, we compute the standard deviation $\hat{\sigma}(x)$ of the tree-level probabilities and define a simple interval around the mean prediction:
\[
C(x) = [ \max\{0, \hat{p}(x) - \hat{\sigma}(x)\},\; \min \{1, \hat{p}(x) + \hat{\sigma}(x)\}]
\]
This approach offers an intuitive way to visualize prediction variability within the ensemble. However, these intervals do not provide any formal statistical guarantees of coverage. In particular, they rely on the assumption that the distribution of tree predictions is roughly symmetric and unimodal, and the constructed intervals may become overly wide and unreliable when tree predictions are highly variable or skewed. A more robust and practically useful alternative would offer theoretical coverage guarantees while producing tighter, more informative intervals for decision-making.

\subsection{Conformal-Calibrated Intervals (CCI)}
To obtain statistically valid confidence prediction intervals around probabilistic outputs in binary classification, we propose an adaptive Conformal-Calibrated Intervals (CCI) approach. The key idea is to calibrate a pre-trained probabilistic classifier (e.g., logistic regression or neural network) using a nonconformity score function and a held-out calibration set to quantify uncertainty in the predicted probabilities. Given a classifier \( f: \mathcal{X} \to [0,1] \) that outputs predicted probabilities \( p = f(x) \), we first map true binary labels \( y \in \{0,1\} \) to continuous centers within the unit interval:

\begin{equation}
\label{eqn:y_scaled}
    y' = 0.25 + 0.5 \cdot y
\end{equation}

This scaling creates distinct interval centers corresponding to each class, which facilitates interval construction in the probability space.

To account for the varying uncertainty inherent in probability predictions, we define an adaptive uncertainty scaling function \(\sigma: [0,1] \to \mathbb{R}^+\) as:

\begin{equation}
    \sigma(p) = 1 + (1 - |p - 0.5|)
\end{equation}

This linear function increases the uncertainty scaling as the predicted probability ($p$) approaches the decision threshold at 0.5, where the classifier is least confident, and decreases it toward the extremes (0,1), where predictions are more certain -- increasing interpretability while effectively modeling the inherent ambiguity in probability estimates.  

Given a calibration set \( \{(x_i, y_i)\}_{i=1}^{n} \), we compute predicted probabilities \( p_i = f(x_i) \) and nonconformity scores:

\begin{equation}
    S(p_i, y'_i) = \frac{\left(y'_i - p_i\right)^2}{\sigma(p_i)}
\end{equation}

The score $S(p_i, y'_i)$ is well-suited for binary classification because it uses the squared error between the transformed label $y'_i$ and the predicted probability $p'_i$, normalized by the uncertainty scaling $\sigma(p_i)$. It penalizes larger deviations in regions of high uncertainty (near $p = 0.5$) while being more lenient for confident predictions (near 0 or 1). This design explicitly encodes the clinical intuition that borderline predictions should be treated more cautiously, while still preserving conformal calibration at the marginal level.

\begin{equation}
    \hat{q} = \min \left\{ q \in \mathbb{R} : \frac{1}{n} \sum_{i=1}^{n} \mathbf{1}[S(p_i, y'_i) \leq q] \geq 1 - \alpha \right\}
\end{equation}

For a input \( x_{\mathrm{test}} \) with predicted probability \( \hat{p} = f(x_{\mathrm{test}}) \), the interval center is chosen based on the predicted class as:

\[
y'_{\mathrm{test}} = 0.75 - 0.5 \cdot \mathbf{1}[\hat{p} < 0.5]
\]
assigning the closest class center to the predicted probability, consistent with the calibration transformation.

Finally, the Conformal-Calibrated interval is defined as:

\begin{equation}
\label{eqn:cp_interval}
    C(x_{\mathrm{test}}) = \left\{ p_{\mathrm{test}} \in [0,1] \;\middle|\; S(p_{\mathrm{test}}, y'_{\mathrm{test}})\leq \hat{q} \right\}    
\end{equation}

This approach yields prediction intervals that adapt in width to the model's estimated uncertainty. Unlike naive variance-based intervals, our CCI method ensures valid marginal coverage under exchangeability, while maintaining simplicity and interpretability through the choice of the linear uncertainty function \(\sigma\).

\begin{figure}[h]
    \centering
    \includegraphics[height=0.1\textheight]{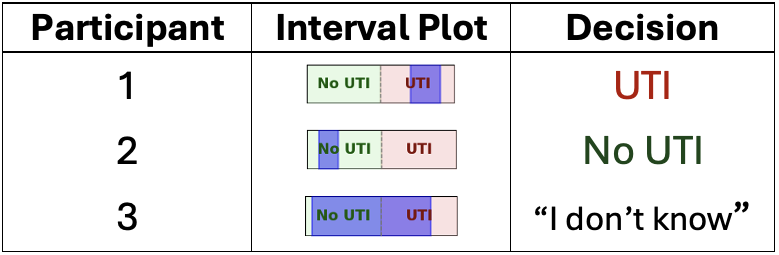}
    \caption{Illustration of interval-based decisions for three participants. The purple bands represent predicted intervals.}
    \label{fig:decision_plot}
\end{figure}

\vspace{0.75ex}
THEOREM 1. { \em Assuming that the calibration and test data are exchangeable, the prediction interval $C(X_{\mathrm{test}})$ for a new testing input $X_{\mathrm{test}}$ is guaranteed to contain the transformed label $y'_{test}$ with a probability of at least $1-\alpha$, where $\alpha$ is the user-specified error rate to configure CCI.}

\begin{equation}\label{eq:mtcp_coverage_theorem}
    \mathbb{P}[Y'_{\text{test}} \in C(X_{\text{test}}) ] \geq 1 - \alpha
\end{equation}

Proof is provided in the Appendix.

\section{Experiments and Results}
\subsection{Experimental Setup}

\subsubsection{Datasets.}
We conducted our study using data collected from the Center for Advanced Studies in Adaptive Systems (CASAS) smart homes \cite{cook2012casas}. These datasets were collected via Internet of Things (IoT) sensors unobtrusively deployed throughout participants’ homes to monitor daily activities. The sensor network included passive infrared motion detectors placed in each functional area, including bedrooms, bathrooms, living rooms, dining rooms, and kitchens. Additional sensors were placed above beds, sinks, toilets, and entryways. Door sensors were installed on external doors and cabinets containing medicine. Motion detectors were coupled with ambient light sensors, and door sensors were coupled with ambient temperature sensors. 

After preprocessing and filtering, the final dataset comprised 117 labeled daily data points. Table~\ref{table:participant_demographics} summarizes participant demographics.

\subsubsection{Models.} We formulated the UTI detection task as a binary classification problem, where the goal is to predict whether a UTI event occurred on a given day based on behavioral features extracted from smart home sensor data. We evaluated two widely used ML models (Logistic Regression and a Neural Network) and one random baseline. Model training procedure and hyperparameter details are in the Appendix.

\subsubsection{UQ Methods.} To quantify uncertainty in the predicted probabilities, we applied two uncertainty quantification methods to the base models. 1) A naive approach, utilizing Random Forest to estimate variance across predictions, constructed confidence intervals based on the ensemble’s variability. 2) Our proposed method, the Conformal Calibration Interval (CCI), operates on a trained ML classifier to provide statistically guaranteed coverage, which is critical for reliable medical decision-making. For these experiments, we partitioned the dataset by allocating 10\% of the entire data for testing, 40\% of the remaining for calibration, and the rest for training. We set the error rate $\alpha =0.1$, and all results are averaged over 20 independent runs to ensure robustness.

\subsubsection{Interval-based Decision.} To translate uncertainty intervals into actionable decisions, we adopt an interval-based rule that classifies each prediction into one of three outcomes: \{``UTI", ``No UTI", ``I don't know"\}. Specifically, if the prediction interval or the associated right-tail probability ($p_{\text{right}}$) strongly supports a positive case (i.e., lower bound $\geq 0.5$ or $p_{\text{right}} \geq 1-\alpha$), the decision is \textbf{``UTI"}. Conversely, if the interval or left-tail probability ($p_{\text{left}}$) indicates strong confidence in the negative class (i.e., upper bound $< 0.5$ or $p_{\text{left}} \geq 1-\alpha$), the decision is \textbf{``No UTI"}. When the prediction interval overlaps the decision boundary and neither tail shows strong certainty, the model abstains with the label \textbf{``I don't know''}. This abstention policy flags ambiguous cases for further nurse evaluation, which is critical for avoiding misdiagnosis in highly uncertain cases. 

\subsubsection{Evaluation Metrics.} To assess model performance, we evaluated all methods—both interval-based (Naive and Conformal-Calibrated) and non-interval-based (Logistic Regression and Random Guess)—using standard classification metrics of accuracy, precision, recall, and F1 score.

For interval-based models that include the option to abstain, we conducted evaluations in two complementary ways. First, to ensure fair comparison with non-abstaining baselines, we report accuracy, precision, recall, and F1 only on non-abstained predictions (i.e., ``UTI" or ``No UTI" decisions). Second, we evaluated abstention behavior using two key metrics: (1) abstention proportion and (2) interval width, as frequent abstentions or excessively wide intervals reduce clinical utility by limiting actionable decisions or providing less informative uncertainty estimates. 

\begin{table*}[ht]
    \renewcommand{\arraystretch}{1.2}
    \begin{tabular}{|c|cccc|ccc|}
    \hline
    
     &  & &  &  & \textbf{Abstention}  & \textbf{Interval}  & \textbf{Interval}  \\
    
     \textbf{Method}  & \textbf{Accuracy} & \textbf{Precision} & \textbf{Recall} & \textbf{F1} &  \textbf{Proportion}  &  \textbf{Width} &  \textbf{plot} \\
    \hline

    Random Guess   & 0.49 $\pm$ 0.14  &  0.48 $\pm$ 0.29  &  0.24 $\pm$ 0.14  &  0.32 $\pm$ 0.19  & \textbf{--} & \textbf{--} & \textbf{--} \\
    Base ML Model     & 0.69 $\pm$ 0.15  & 0.68 $\pm$ 0.15  &  0.77 $\pm$ 0.15  &  0.72 $\pm$ 0.12  & \textbf{--} & \textbf{--} & \textbf{--} \\
    \rowcolor[gray]{0.9}
    Naive-interval & 0.71 $\pm$ 0.26  &  0.60 $\pm$ 0.41  &  0.61 $\pm$ 0.40  &  0.57 $\pm$ 0.36  & 0.73 $\pm$ 0.12 & 0.60 $\pm$ 0.06 & \raisebox{-0.3\height}{\includegraphics[width=2cm, height=0.42cm]{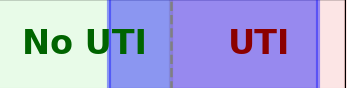}} \\
    \rowcolor[gray]{0.9}
    CCI & \textbf{0.72 $\pm$ 0.16} & \textbf{0.74 $\pm$ 0.17} & \textbf{0.78 $\pm$ 0.17} & \textbf{0.75 $\pm$ 0.14}  & \textbf{0.22 $\pm$ 0.14} & \textbf{0.20 $\pm$ 0.05}  & \raisebox{-0.3\height}{\includegraphics[width=2cm, height=0.42cm]{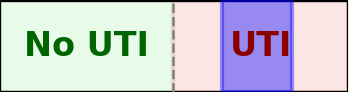}}\\
    
    \hline
    \end{tabular}
    \centering
    \caption{Performance comparison of non-interval-based and interval-based methods (grey rows) over 20 independent runs. Classification metrics (accuracy, precision, recall, F1) are reported for all models, while abstention proportion and prediction interval width are included only for interval-based approaches. Visuals of interval quality are shown in the final column.}
    \label{table:performance_comparison}
\end{table*}

\subsection{Results and Discussion}
\subsubsection{Performance of Baseline Models}
Figure~\ref{fig:base_models_cm2} shows confusion matrices for three non-interval-based methods: Logistic Regression, Neural Network, and Random Guess. Among these, both Logistic Regression and the Neural Network achieve balanced performance, correctly identifying both UTI and No UTI cases. In contrast, the Random Guess model performs poorly, with high misclassification rates across both classes, as expected. As shown in Table~\ref{table:performance_comparison}, Logistic Regression achieves an accuracy of 0.69 and an F1 score of 0.72, while Random Guess lags behind with 0.49 accuracy and 0.32 F1 score. The recall metric, which is critical in UTI detection, is highest for Logistic Regression (0.68), showing its reliability in capturing positive cases.

\subsubsection{Interval-based Methods: Naive vs. Conformal} 
Tables \ref{table:performance_comparison} and \ref{tab:performance_comparison_all_runs} compare interval-based methods in terms of both classification metrics and uncertainty measures. The Naive interval method performs poorly, with an F1 score of just 0.57 and an extremely high abstention rate (0.73), meaning it avoids making predictions most of the time. Its wide average prediction interval (0.60) indicates low confidence even when predictions are made, making it clinically impractical.

The CCI approach not only outperforms the Naive method but also surpasses all other baselines in classification metrics: accuracy (0.72), precision (0.74), recall (0.78), and F1 score (0.75). Crucially, CCI achieves these gains while abstaining far less (0.22 vs.\ 0.73) and producing much tighter prediction intervals (0.20 vs.\ 0.60). These two factors—low abstention and narrow intervals—are particularly important in clinical decision-making, as they maximize the number of actionable predictions while maintaining high confidence, all without sacrificing classification performance.

\subsubsection{Visual Analysis of Interval Quality} The ``Interval Plot" column of Figure \ref{fig:decision_plot}, Tables~\ref{table:performance_comparison} and \ref{tab:performance_comparison_all_runs} visualizes prediction intervals (on a $[0,1]$) interval range for Naive and CCI methods,  with intervals plotted by randomly selecting a lower bound and constructing to reflect the average interval width over 20 runs. Color-coded regions indicate model certainty: green for ``No UTI," red for ``UTI," and purple for the prediction interval. CCI intervals display narrower purple bands (mean width of 0.22), reflecting higher certainty, while Naive intervals are wider (mean width of 0.73) and thus more ambiguous. This tighter interval width in CCI highlights its superior ability to deliver decisive and clinically useful predictions.

\subsubsection{Clinical Relevance.} From a deployment perspective, failing to detect a true UTI can lead to serious complications, and alerting too often can result in clinicians ignoring them. In addition, trusting the prediction is key if clinicians are to rely on the information for clinical decision-making. The CCI model's superior recall, low abstention, and compact prediction intervals suggest it is well-suited for clinical applications, offering reliable predictions and reserving abstentions mostly for genuinely ambiguous cases. This contrasts with the Naive model, which abstains too often and fails to deliver actionable decisions.

\subsection{Clinicians' Feedback on Uncertainty Outputs}
To complement our quantitative evaluation, we surveyed 42 nurses to evaluate the usefulness of our method - whether the visual outputs of our prediction intervals are interpretable and clinically useful. They answered two open-ended questions and six Likert-scale items (``Strongly Disagree''– ``Strongly Agree'') for model prediction, naïve interval, and calibrated conformal interval (CCI) plots.

For \textbf{Question 1} --- ``Would these graphs help you with clinical decision-making, would you act on this information, and how would you decide?'' --- one respondent on CCI: 

\begin{quote}
    \emph{Given more information on the accuracy yes, I think this could help in clinical decision making. I am a non-prescribing medical professional but imagine people wanting a urinalysis to confirm before prescribing antibiotics. Could help decide whether ordering a urinalysis would be indicated.}
\end{quote}
This supports our use case: smart home markers can prompt timely diagnostic actions like ordering confirmatory tests.

\begin{quote}
    Another wrote: \emph{ These graphs would be a good basis for decision making, however it would be nice to see the factors that led to the algorithm's conclusion.}
\end{quote}
This validates our integration of SHAP explanations to reveal feature contributions.

For \textbf{Question 2} --- ``Suggestions for improving graphs, information, and prediction confidence indicators'' 
\begin{quote}
    --- one comment stated: \emph{I would like to see the blue indicator to be far left or right if it's 100\% certainty instead of having it in the middle or to the side.}
\end{quote}
This aligns with our goal of producing clear, well-calibrated intervals.

\begin{quote}
    Another noted: \emph{If the confidence range was smaller, I would be more inclined to use this as a diagnostic tool.}
\end{quote}
This supports our method’s advantage in producing tighter intervals than baselines.

\begin{quote}
   Finally: \emph{Would like to see outcomes data. Feel more comfortable using a tool if evidence of successful probabilities can be shown.}
\end{quote}
This reinforces our focus on theoretically-valid guarantees via conformal prediction.

\vspace{1ex}

\noindent \textbf{Summary.} Overall, feedback confirms that (1) our behavior markers are clinically meaningful, (2) CCI intervals are interpretable, and (3) their smaller widths and lower abstention rates improve practical utility. For the Likert questions,  Nurses gave \textit{mixed feedback} on base model prediction and \textit{negative responses} to naïve intervals, citing a lack of clarity and limited clinical value. In contrast, the calibrated conformal interval (CCI) plots were consistently rated as clear, trustworthy, and useful, making them more favorable for real-world deployment. Details of the survey questions and Likert plots are provided in the Appendix.

\subsection{Practical Deployment Considerations}
The clinician's feedback reinforces the three central contributions of this work. First, our results demonstrate that clinically relevant behaviors associated with UTIs can be detected from the CASAS smart home dataset by defining and extracting behavior markers. Second, we show that prediction intervals, derived from our uncertainty quantification framework, are not only statistically valid but also interpretable in practice. Third, compared to baseline methods, our intervals achieve smaller widths, higher classification metrics, and lower abstention rates, increasing their utility for real-world decision-making. The survey responses highlight that clinicians value transparency in feature importance (addressed through SHAP explanations), prefer tight and clear confidence intervals (addressed through our conformal prediction approach), and desire outcome validation before routine adoption. These findings provide a concrete roadmap for refining our system for clinician-in-the-loop deployment.

\section{Path to Deployment}
The deployment of our clinician-in-the-loop (CIL) smart home system for UTI detection in real-world clinical settings would involve several key steps to ensure reliability, scalability, and seamless integration into healthcare workflows. First, the CIL detection module would be integrated with existing smart home infrastructures, such as CASAS or similar sensor networks, to ensure continuous and secure data collection. Next, automated real-time pipelines would be implemented to extract daily behavioral markers and compute prediction intervals daily. A secure, user-friendly, clinician-facing dashboard would then be developed to visualize daily risk scores, uncertainty intervals, and abstention flags, allowing healthcare practitioners to review and confirm real-time alerts easily. To ensure smooth adoption, the system would ideally be integrated into the electronic health record (EHR) and aligned with clinical workflows. Importantly,  collaboration between the smart home team and healthcare providers will be critical to establishing intervention thresholds, escalation procedures, and response protocols for flagged cases. Finally, multi-month pilot studies would be conducted to gather clinician feedback, enabling refinement of model parameters and interval settings, along with periodic retraining to adapt to changes in residents’ routines.
Pilot studies with home health and palliative care agencies will provide real-world feedback, enabling iterative refinements to algorithms, interfaces, and alert thresholds based on clinician input and patient outcomes. A cost-benefit analysis evaluating reductions in hospitalization and emergency care rates and healthcare costs due to early UTI detection will support broader adoption, positioning the CIL system as a transformative tool for enhancing clinician-guided care for older adults with chronic conditions.

\section{Conclusion}
We presented a clinician-in-the-loop (CIL) smart home system for detecting urinary tract infection (UTI) flare-ups in older adults, leveraging ambient sensor data and a novel Conformal-Calibrated Interval (CCI) method for uncertainty quantification. By extracting clinically relevant behavioral markers, our system achieves a robust UTI detection policy that outperforms baseline methods in terms of accuracy, precision, recall, F1-score, compact prediction intervals, and a low abstention proportion; enhancing both model interpretability and trust, which leads to improved clinical usefulness. Feedback from 42 nurses confirms that interval-based outputs support informed decision-making in practice. These findings highlight our model's potential for uncertainty-aware smart home monitoring to improve timely interventions and health outcomes, paving the way for larger-scale deployment in remote health monitoring.

\section{Acknowledgements}
Research reported in this publication was supported by the National Institutes of Nursing Research of the National Institutes of Health under award number R01NR016732 and by the National Science Foundation under award number 1854362. The content is solely the responsibility of the authors and does not necessarily represent the official views of the National Institutes of Health or the National Science Foundation.

\bigskip

\bibliography{aaai2026}
\newpage
\input{appendix}

\end{document}

%% file: appendix.tex
\clearpage
\onecolumn
\section{Appendix}
\subsection{Proof}

\noindent \textbf{Theorem 1 Proof.} \textit{(Marginal coverage of the conformal–calibrated predictor)}

Assume the calibration sample $\{(X_i,Y_i)\}_{i=1}^n$ and the test point $(X_{\text{test}},Y_{\text{test}})$ are exchangeable.
Let $f:\mathcal{X}\to[0,1]$ be a fixed pre–trained probabilistic classifier and define
$Y_i' = 0.25 + 0.5\,Y_i$ and $p_i=f(X_i)$.  
With the nonconformity score function 
\[
S(p,y') = \frac{(y'-p)^2}{\sigma(p)},
\]
define the calibration scores $s_i = S(p_i,Y_i')$, $i=1,\dots,n$, and let 
\[
\hat q
=\min\Bigl\{q\in\mathbb{R}:\frac{1}{n}\sum_{i=1}^n\mathbf{1}\{s_i\le q\}\ge 1-\alpha\Bigr\}.
\]
For a new feature vector $x_{\mathrm{test}}$ with predicted class-specific center $y'_{\mathrm{test}}$, the conformal prediction set is
\begin{equation}
\label{eq:label_set}
    C(x_{\mathrm{test}}) = \left\{ p \in [0,1] \;\middle|\; S(p, y'_{\mathrm{test}})\leq \hat{q} \right\}.
\end{equation}

Let 
\[
s_{n+1} = S(f(X_{\mathrm{test}}),Y'_{\text{test}})
\]
denote the (unknown) test score.  
By construction, the $n+1$ random variables
\[
s_1,\ldots,s_n,s_{n+1}
\]
are exchangeable. Hence,
\[
\mathbb{P}\{\, s_{n+1} \le s_{(k)} \,\} \;\ge\; \frac{k}{n+1} \;\ge\; 1-\alpha .
\]
Next observe that $\hat q$, as defined above from the $n$ calibration scores,
satisfies $s_{(k)}\ge \hat q$ (the usual property that the empirical
$(1-\alpha)$ quantile from $n$ points is no larger than the corresponding order statistic from $n+1$ points). Therefore
\[
\mathbb{P}\{\, s_{n+1} \le \hat q \,\}
\;\ge\; \mathbb{P}\{\, s_{n+1} \le s_{(k)} \,\}
\;\ge\; 1-\alpha .
\]
Finally, by the definition of $C(\cdot)$ in \eqref{eq:label_set},
\[
S\bigl( f(X_{\mathrm{test}}),Y'_{\text{test}}\bigr)\le \hat q
\quad\Longleftrightarrow\quad
Y'_{\text{test}} \in C(X_{\text{test}}),
\]
which yields
\begin{equation}\label{eq:mtcp_coverage_theorem_1}
    \mathbb{P}[Y'_{\text{test}} \in C(X_{\text{test}}) ] \geq 1 - \alpha .
\end{equation}

\textit{\textbf{Remark.} The prediction set $C(x_{\mathrm{test}})$ is visualized as an interval in probability space
centered at the class-specific anchors $\{0.25,0.75\}$.
This is equivalent to \eqref{eq:label_set}: it is precisely the set of probabilities whose
score with label center $y'$ does not exceed $\hat q$. Thus
$Y'_{\text{test}}\in C(X_{\text{test}})$ iff
$S( f(X_{\mathrm{test}}),Y'_{\text{test}})\le \hat q$, and the marginal
coverage $1-\alpha$ follows directly.}

\subsection{Detailed Behavior features}

From the continuous ambient smart home sensor data, we extracted 17 clinically relevant features that reflected daily behavior. These features could be fed into a machine learning algorithm to classify each day as indicating the event of a UTI flare-up or not. The featrues are grouped into the following categories, with complete detailes presentedin the appendix:

\begin{itemize}

    \item \textit{Daily Bathroom Visit Count.} This feature represents the total number of distinct bathroom visits recorded during a day. A visit is identified when a series of ON events from bathroom sensors is followed by at least 5 minutes of inactivity before another sequence begins, indicating a new visit.
    
    \item \textit{Average Bathroom Visit Duration (min).} For each detected bathroom visit, the duration is calculated as the time span from the first to the last ON event in the visit. The average duration across all visits in a day is reported in minutes, reflecting the typical time spent per visit.
    
    \item \textit{Nocturnal Bathroom Visits.} This captures the number of bathroom visits occurring during nocturnal hours (9pm – 7am). Visits are extracted using the same gap-based method, applied only to events within the defined nocturnal window.
    
    \item \textit{Mean Bed-to-Bathroom Transit Time (s).} This feature measures the average time (in seconds) taken to transition from a bedroom event (e.g., bed exit) to a bathroom event within a 5-minute window. It provides insight into nighttime mobility and response time.
    
    \item \textit{Standard Deviation of Transit Time (s).} The standard deviation of bedroom-to-bathroom transit times over a day, reflecting variability in how quickly individuals move between these locations. High variability could indicate inconsistent mobility patterns.
    
    \item \textit{Mean Daytime Transit Time (s).} Similar to Mean Transit Time, but restricted to non-nocturnal hours (7 AM–9 PM). This distinguishes daytime mobility characteristics from nighttime patterns.
    
    \item \textit{Standard Deviation of Daytime Transit Time (s).} This feature captures the variability in bed-to-bathroom transitions during the day, offering additional insight into consistency in activity during waking hours.
    
    \item \textit{Daily Movement Entropy.} Computed as the Shannon entropy of ON events across all home sensors, this feature quantifies the diversity and unpredictability of movement throughout the home. Higher entropy suggests more dispersed and varied activity.
    
    \item \textit{Nocturnal Awakenings (Sensor-Based).} Counts the number of times an individual is inferred to be awake during the night, based on OFF events from the bed sensor or ON events from any non-bed sensor. It indicates restlessness or disrupted sleep.
    
    \item \textit{Early Morning Awakenings.} This feature counts distinct episodes of sensor activity, indicating awakenings before 6:00 AM. Clusters of activity separated by at least 10 minutes are considered separate awakenings.
    
    \item \textit{Consecutive Bathroom Activity Episodes.} Identifies clusters of bathroom activity where multiple sensors are triggered within a 30-minute window. These clusters may reflect repeated or prolonged use of the bathroom within short time spans.
    
    \item \textit{Nocturnal Non-Bathroom Movement.} The count of ON events from non-bathroom sensors between 10 PM and 7 AM, representing activity in other areas of the home during hours typically associated with rest.

    \item \textit{Health Event (Last 3 days)} A retrospective indicator set to 1 if any relevant health event (e.g., UTI) occurred within the three days preceding the current day, and 0 otherwise. It supports analysis of recent conditions that may precede notable changes in behavior.
    
    \item \textit{Change in Bathroom Visit Frequency.} The day-to-day difference in bathroom visit frequency, computed as the current day’s count minus the previous day’s count. This feature captures abrupt increases or decreases in usage.
    
    \item \textit{3-Day Rolling Std. of Visit Duration.} Calculates the standard deviation of average bathroom visit durations over a 3-day sliding window. This temporal smoothness highlights fluctuations in bathroom behavior.
    
    \item \textit{Percent of Visits at Night.} Computes the percentage of bathroom visits that occur at night. This metric contextualizes nighttime usage relative to overall behavior.
    
    \item \textit{3-Day Rolling Avg. Bathroom Visits.} The mean of daily bathroom visit frequencies over a 3-day window. This smooths short-term fluctuations and helps identify sustained changes in behavior.

\end{itemize}

\begin{figure}[h]
    \centering
    \includegraphics[width=1\linewidth]{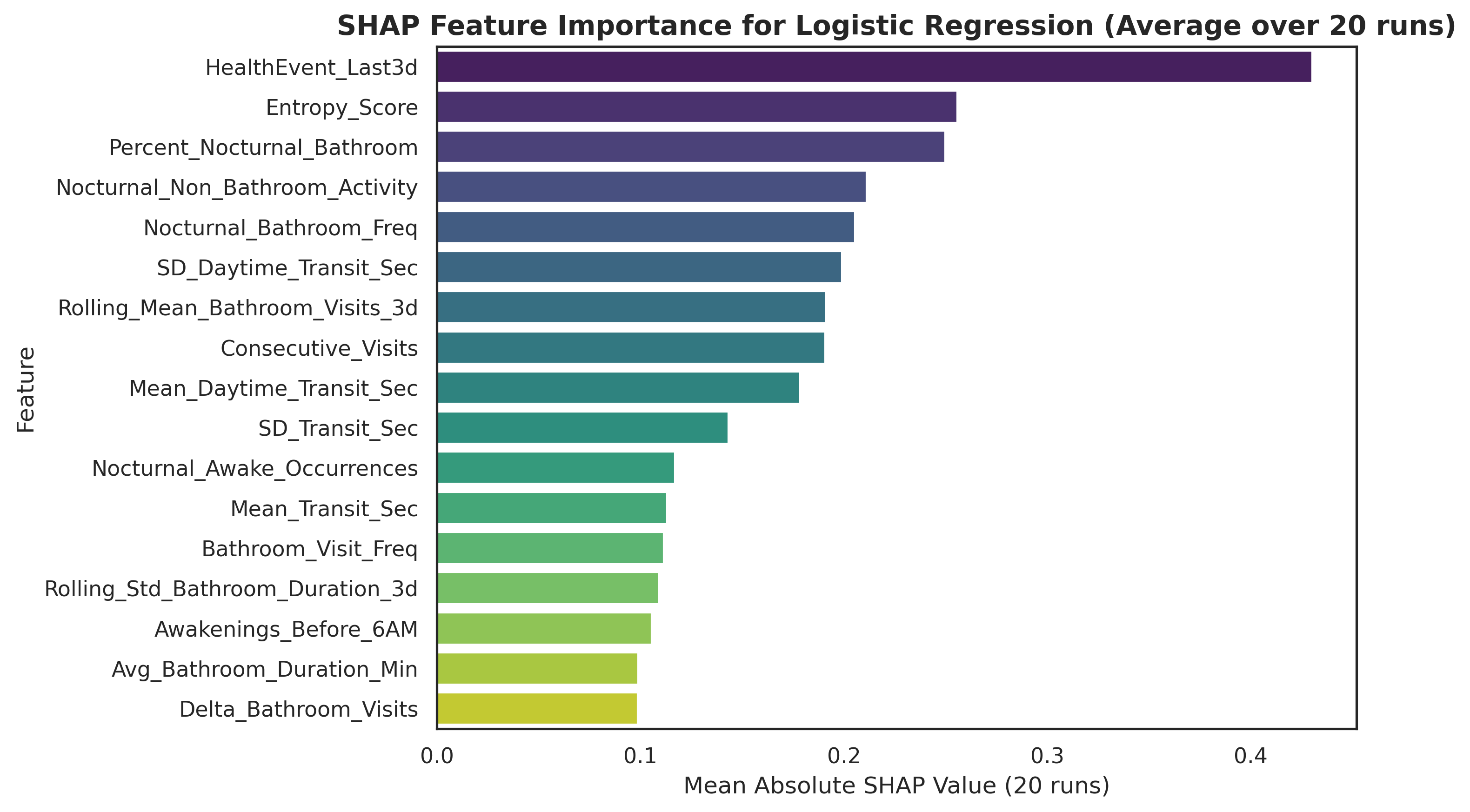}
    \caption{SHAP Feature importance for all extracted features for the base model (averaged over 20 independent runs)}
    \label{fig:SHAP}
\end{figure}

\subsection{Models}
\begin{figure}[ht]
    \centering
    \includegraphics[width=0.9\linewidth]{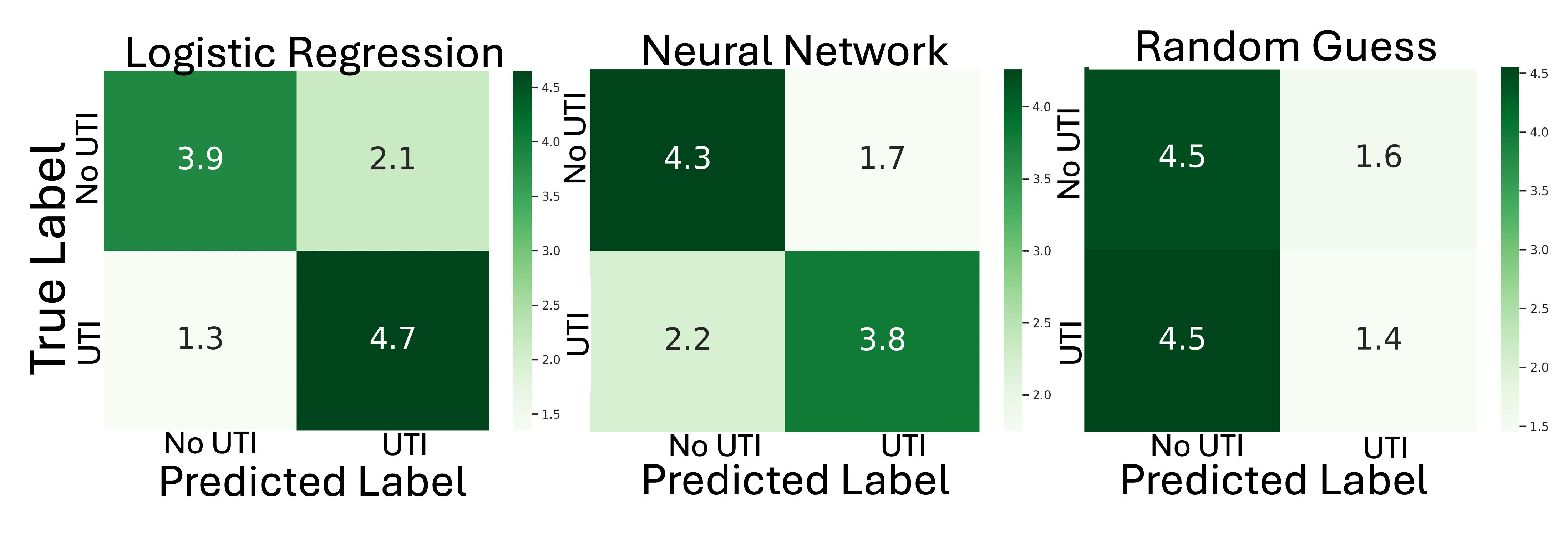}
    \caption{Confusion matrices for Logistic Regression, Neural Network, and Random Guess. Values are averaged over 20 runs.}
    \label{fig:base_models_cm2}
\end{figure}

We evaluated two widely used models selected for their simplicity: Logistic Regression and a Neural Network. Logistic Regression was chosen for its interpretability and ability to capture linear relationships in sensor data. The Neural Network was selected to model non-linear patterns, such as irregular activity sequences indicating UTI flare-ups. Hyperparameter tuning was performed using GridSearchCV with 3-fold cross-validation over the 117 data points, using F1 scoring. For Logistic Regression, tuned parameters included C (regularization strength) and solver, with optimal values typically C = 0.1 and solver = `liblinear' across 20 runs. For the Neural Network, tuned parameters included alpha (regularization term), hidden layer sizes, and learning rate, with optimal values typically alpha = 0.0001, hidden layer sizes = (50,), and learning rate = `constant' across 20 runs. Logistic Regression was selected as the primary model due to its consistent superiority over the Neural Network across multiple metrics (accuracy, precision, recall, and F1), as detailed in Figure \ref{fig:base_models_cm2} and the Evaluation Metrics Section. We benchmarked our models against a Random Guess baseline, which assigned labels randomly and performed significantly worse, highlighting the robustness of our trained models.

\subsubsection{Machine Specifications:} All experiments were conducted on the following hardware and software setup:

\begin{itemize}
    \item \textbf{Operating System:} Rocky Linux 8.10 (Green Obsidian), \textbf{Processor:} AMD EPYC 7573X 32-Core Processor
    \item \textbf{GPUs:} Two NVIDIA A40 GPUs (each with 46 GB of memory), \textbf{GPU Driver Version:} 555.42.02, \textbf{CUDA Version:} 12.5
\end{itemize}



\section{Survey: Nurses' Perceptions of Plot Usefulness}

To evaluate the usefulness of our method, 42 nurses were asked to respond to six questions using a seven-point Likert scale ranging from ``Strongly Disagree'' to ``Strongly Agree.'' The questions assessed whether the plots (1) clearly convey the likelihood of urinary tract infection (UTI), (2) instill confidence in supporting a diagnosis, (3) present information that can be trusted, (4) could influence clinical decision-making, (5) help in communicating with caregivers or patients, and (6) would be considered for use in clinical workflow. Responses were collected separately for the point prediction plots, naïve interval plots, and the calibrated conformal interval (CCI) plots.

\begin{figure}[ht]
    \centering
    \includegraphics[width=1\linewidth]{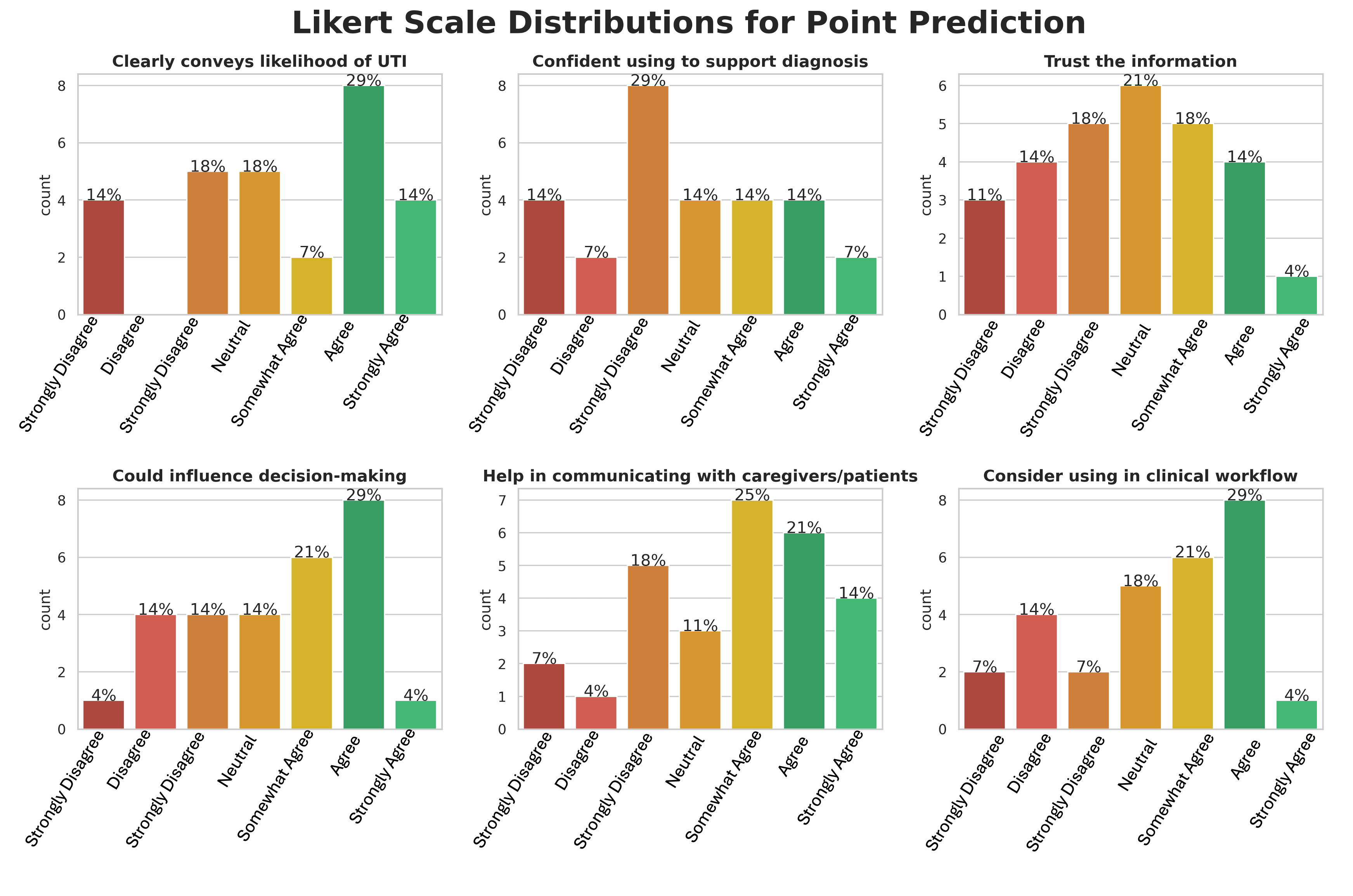}
    \caption{Survey responses distribution of Likert scale responses for base model (point prediction) across six dimensions: clarity in conveying UTI likelihood, confidence in using to support diagnosis, trust in the information, potential influence on decision-making, help in communicating with caregivers/patients, and consideration for use in clinical workflow. Responses for the Naive Interval show a wider spread with higher disagreement, while those for CCI are predominantly positive with strong agreement.}

    \label{fig:point_pred_likert}
\end{figure}

\begin{figure}
    \centering
    \includegraphics[width=\linewidth, height=0.9\textheight, keepaspectratio]{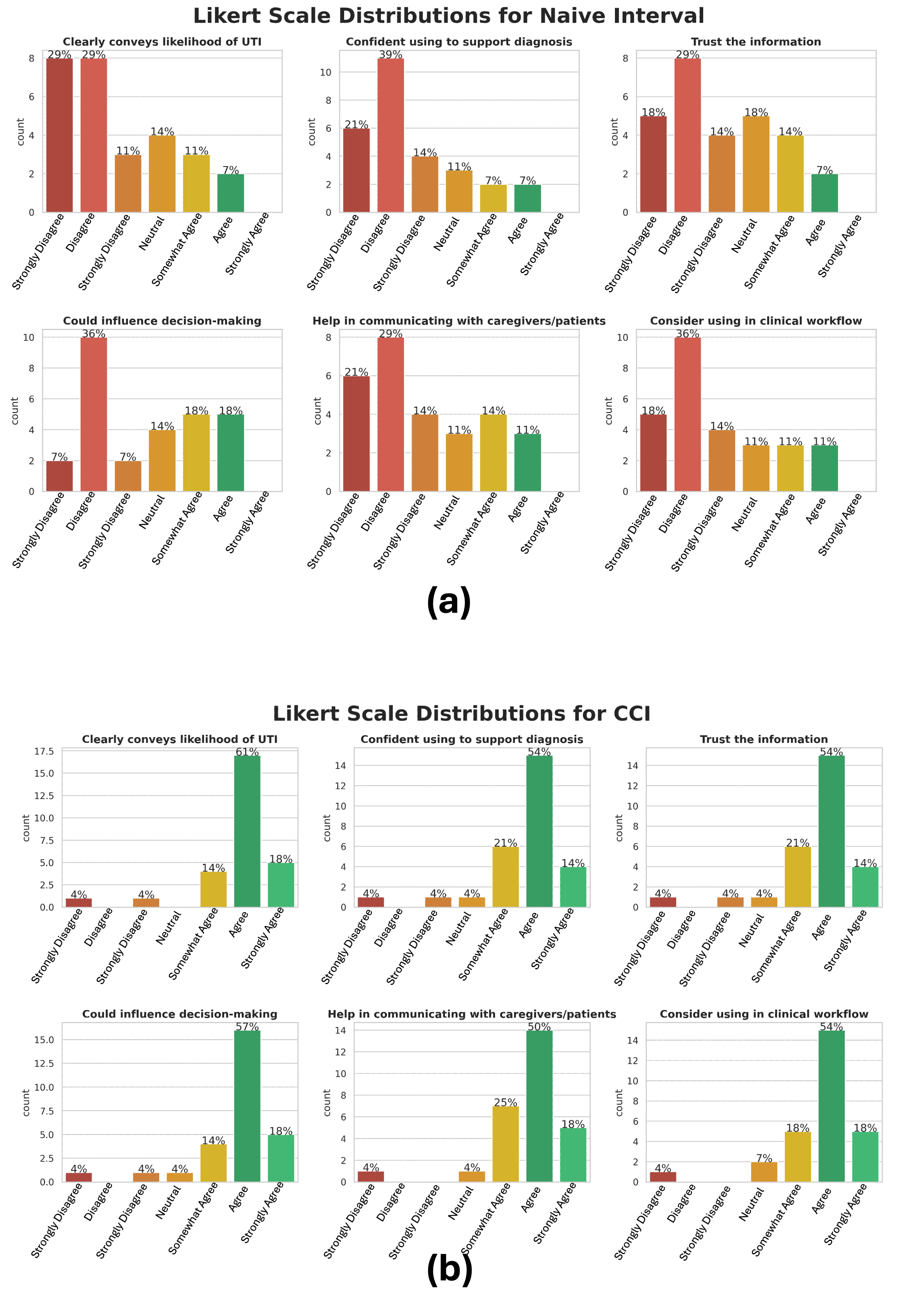}
    \caption{Survey responses distribution of Likert scale responses for the Naive Interval (panel a) and CCI (panel b) across six dimensions: clarity in conveying UTI likelihood, confidence in using to support diagnosis, trust in the information, potential influence on decision-making, help in communicating with caregivers/patients, and consideration for use in clinical workflow. Responses for the Naive Interval show a wider spread with higher disagreement, while those for CCI are predominantly positive with strong agreement.}
    \label{fig:interval_likert}
\end{figure}

The results highlight clear differences between the three approaches. The point prediction plots in Figure \ref{fig:point_pred_likert} show a relatively even split between positive and negative feedback, with responses distributed across the Likert scale: some nurses considered them beneficial, while others voiced doubts or reservations about their clarity and practical application. Conversely, the naïve interval plots (Figure \ref{fig:interval_likert}a) elicited mostly negative responses, with a significant number of nurses disagreeing that these plots were clear, trustworthy, or valuable in clinical settings. Notably, many strongly rejected the idea that these plots effectively communicated UTI likelihood or impacted decision-making, and confidence in their diagnostic use was minimal. On the other hand, the CCI plots garnered highly favorable reactions, with the majority of nurses agreeing or strongly agreeing across all six questions that the CCI plots (Figure \ref{fig:interval_likert}b) were clear, trustworthy, and useful. For instance, more than half of the respondents supported their potential to influence decision-making and suitability for clinical workflow integration.


Overall, these results suggest that while naïve interval plots may create confusion and reduce trust, the CCI approach substantially improves perceived clarity, confidence, and clinical usefulness. Nurses viewed the CCI plots as the most reliable and actionable form of visualization, indicating that calibrated interval-based communication has strong potential to support diagnostic reasoning and integration into clinical practice.

\clearpage

\subsection{Additional Results}

\begin{table*}[ht]
    \centering
    \renewcommand{\arraystretch}{1.2}
    \begin{tabular}{|c|ccc|ccc|}
    \hline
     & \multicolumn{3}{c|}{\textbf{Naive Intervals}} & \multicolumn{3}{c|}{\textbf{CCI}}\\
     {Runs} & Abstention & Interval & Interval & Abstention & Interval & Interval \\
     & proportion & width & Plot & proportion & width & Plot \\
    \hline
    1 & 0.91 & 0.55 & \raisebox{-0.3\height}{\includegraphics[width=2cm, height=0.6cm]{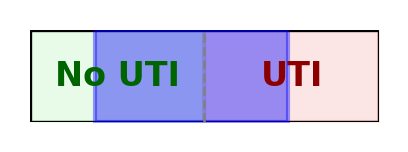}} & \textbf{0.25} & \textbf{0.10} & \raisebox{-0.3\height}{\includegraphics[width=2cm, height=0.6cm]{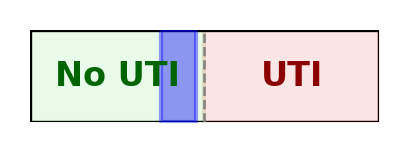}} \\
    2 & 0.67 & 0.58 & \raisebox{-0.3\height}{\includegraphics[width=2cm, height=0.6cm]{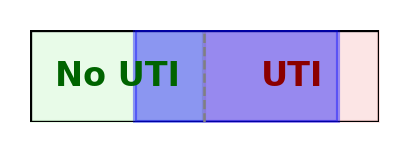}} & \textbf{0.25} & \textbf{0.26} & \raisebox{-0.3\height}{\includegraphics[width=2cm, height=0.6cm]{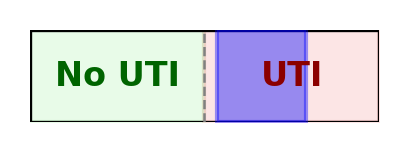}} \\
    3 & 0.75 & 0.64 & \raisebox{-0.3\height}{\includegraphics[width=2cm, height=0.6cm]{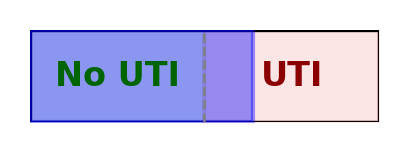}} & \textbf{0.08} & \textbf{0.24} & \raisebox{-0.3\height}{\includegraphics[width=2cm, height=0.6cm]{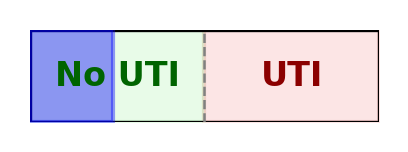}} \\
    4 & 0.58 & 0.55 & \raisebox{-0.3\height}{\includegraphics[width=2cm, height=0.6cm]{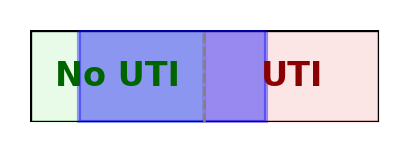}} & \textbf{0.08} & \textbf{0.16} & \raisebox{-0.3\height}{\includegraphics[width=2cm, height=0.6cm]{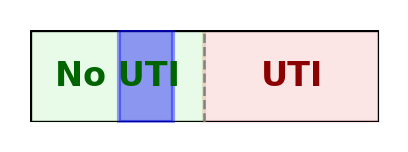}} \\
    5 & 0.67 & 0.59 & \raisebox{-0.3\height}{\includegraphics[width=2cm, height=0.6cm]{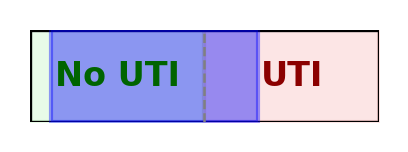}} & \textbf{0.08} & \textbf{0.18} & \raisebox{-0.3\height}{\includegraphics[width=2cm, height=0.6cm]{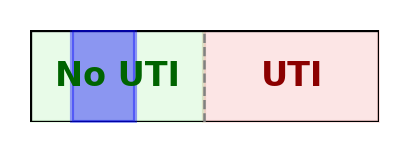}} \\
    6 & 0.83 & 0.69 & \raisebox{-0.3\height}{\includegraphics[width=2cm, height=0.6cm]{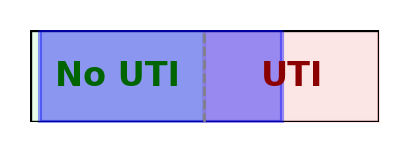}} & \textbf{0.08} & \textbf{0.13} & \raisebox{-0.3\height}{\includegraphics[width=2cm, height=0.6cm]{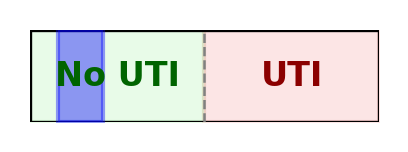}} \\
    7 & 0.67 & 0.70 & \raisebox{-0.3\height}{\includegraphics[width=2cm, height=0.6cm]{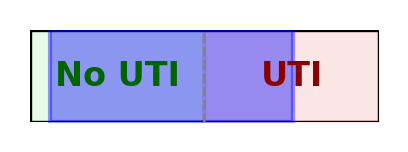}} & \textbf{0.25} & \textbf{0.14} & \raisebox{-0.3\height}{\includegraphics[width=2cm, height=0.6cm]{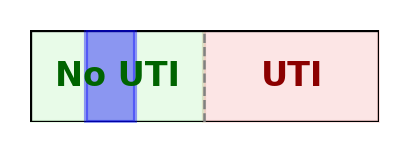}} \\
    8 & 0.75 & 0.55 & \raisebox{-0.3\height}{\includegraphics[width=2cm, height=0.6cm]{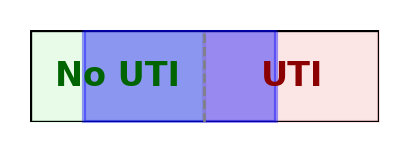}} & \textbf{0.33} & \textbf{0.17} & \raisebox{-0.3\height}{\includegraphics[width=2cm, height=0.6cm]{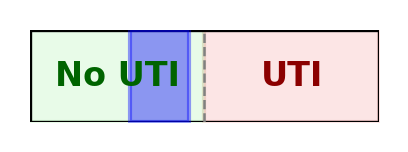}} \\
    9 & 0.67 & 0.62 & \raisebox{-0.3\height}{\includegraphics[width=2cm, height=0.6cm]{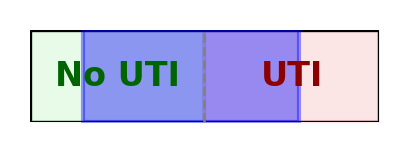}} & \textbf{0.08} & \textbf{0.22} & \raisebox{-0.3\height}{\includegraphics[width=2cm, height=0.6cm]{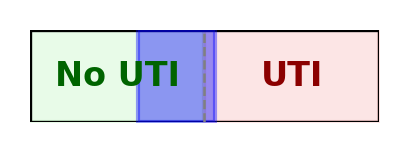}} \\
    10 & 0.67 & 0.63 & \raisebox{-0.3\height}{\includegraphics[width=2cm, height=0.6cm]{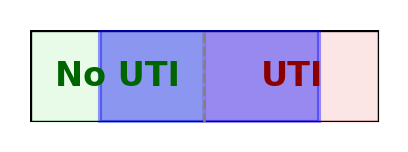}} & \textbf{0.08} & \textbf{0.21} & \raisebox{-0.3\height}{\includegraphics[width=2cm, height=0.6cm]{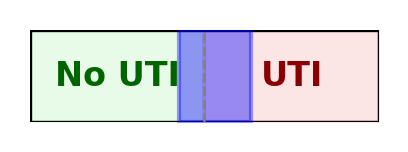}} \\
    11 & 0.83 & 0.62 & \raisebox{-0.3\height}{\includegraphics[width=2cm, height=0.6cm]{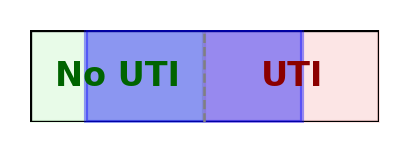}} & \textbf{0.25} & \textbf{0.18} & \raisebox{-0.3\height}{\includegraphics[width=2cm, height=0.6cm]{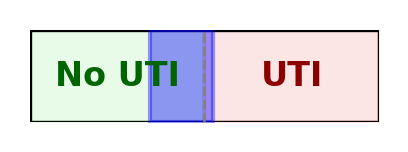}} \\
    12 & 0.67 & 0.56 & \raisebox{-0.3\height}{\includegraphics[width=2cm, height=0.6cm]{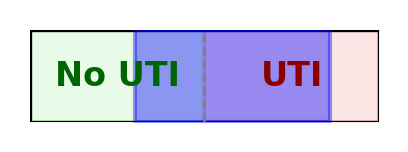}} & \textbf{0.50} & \textbf{0.21} & \raisebox{-0.3\height}{\includegraphics[width=2cm, height=0.6cm]{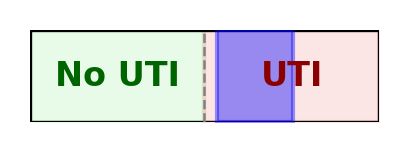}} \\
    13 & 1.00 & 0.78 & \raisebox{-0.3\height}{\includegraphics[width=2cm, height=0.6cm]{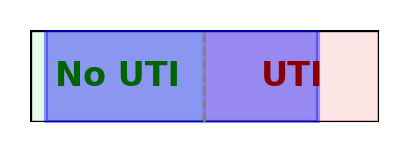}} & \textbf{0.42} & \textbf{0.20} & \raisebox{-0.3\height}{\includegraphics[width=2cm, height=0.6cm]{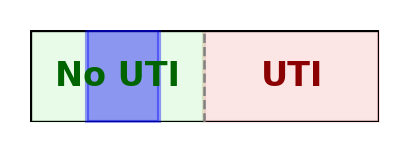}} \\
    14 & 0.67 & 0.57 & \raisebox{-0.3\height}{\includegraphics[width=2cm, height=0.6cm]{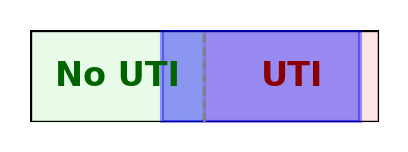}} & \textbf{0.33} & \textbf{0.20} & \raisebox{-0.3\height}{\includegraphics[width=2cm, height=0.6cm]{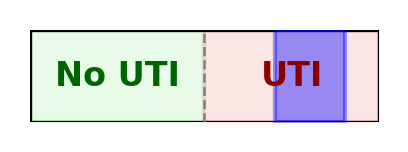}} \\
    15 & 0.92 & 0.57 & \raisebox{-0.3\height}{\includegraphics[width=2cm, height=0.6cm]{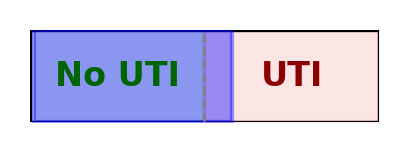}} & \textbf{0.50} & \textbf{0.33} & \raisebox{-0.3\height}{\includegraphics[width=2cm, height=0.6cm]{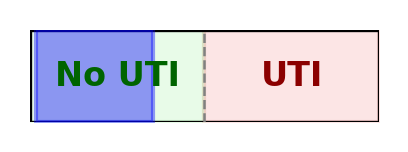}} \\
    16 & 0.75 & 0.58 & \raisebox{-0.3\height}{\includegraphics[width=2cm, height=0.6cm]{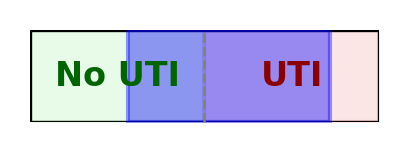}} & \textbf{0.00} & \textbf{0.18} & \raisebox{-0.3\height}{\includegraphics[width=2cm, height=0.6cm]{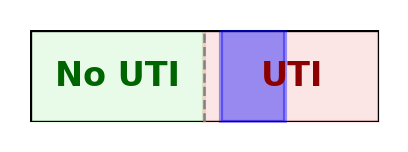}} \\
    17 & 0.67 & 0.53 & \raisebox{-0.3\height}{\includegraphics[width=2cm, height=0.6cm]{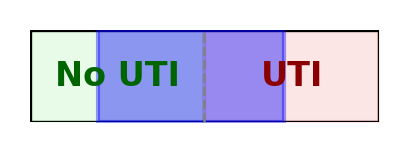}} & \textbf{0.17} & \textbf{0.17} & \raisebox{-0.3\height}{\includegraphics[width=2cm, height=0.6cm]{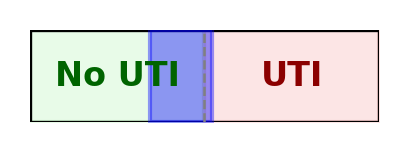}} \\
    18 & 0.58 & 0.55 & \raisebox{-0.3\height}{\includegraphics[width=2cm, height=0.6cm]{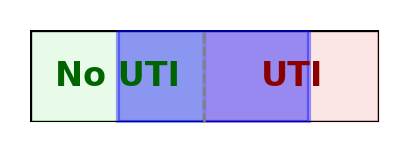}} & \textbf{0.17} & \textbf{0.22} & \raisebox{-0.3\height}{\includegraphics[width=2cm, height=0.6cm]{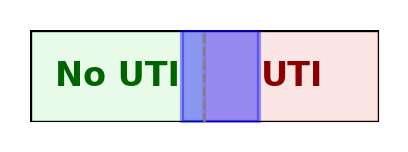}} \\
    19 & 0.59 & 0.53 & \raisebox{-0.3\height}{\includegraphics[width=2cm, height=0.6cm]{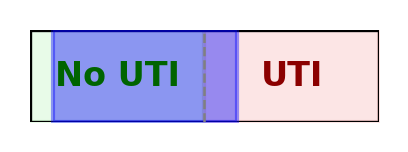}} & \textbf{0.33} & \textbf{0.29} & \raisebox{-0.3\height}{\includegraphics[width=2cm, height=0.6cm]{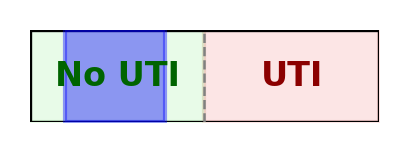}} \\
    20 & 0.67 & 0.61 & \raisebox{-0.3\height}{\includegraphics[width=2cm, height=0.6cm]{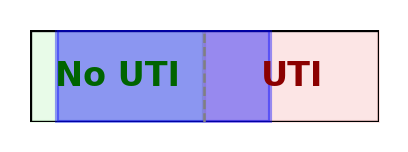}} & \textbf{0.17} & \textbf{0.27} & \raisebox{-0.3\height}{\includegraphics[width=2cm, height=0.6cm]{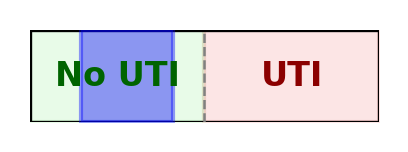}} \\
    \hline
    \end{tabular}
    \caption{Comparison of Performance Metrics across interval-based Methods (20 runs). Naive intervals exhibit a high abstention proportion (0.73) and wider intervals (0.60), whereas CCI performs better with a lower abstention proportion (0.22) and narrower intervals (0.20) - relevant for practical utility.}
    \label{tab:performance_comparison_all_runs}
\end{table*}




